# Automated Romberg Test: Leveraging a CNN and Centre of Mass Analysis for Sensory Ataxia Diagnosis

Reilly Haskins

*Dept. of Computer Science and Software Engineering*
*University of Canterbury*

Christchurch, New Zealand
reilly.haskins@pg.canterbury.ac.nz

Richard Green

*Dept. of Computer Science and Software Engineering*
*University of Canterbury*

Christchurch, New Zealand
richard.green@canterbury.ac.nz

*Abstract*—**This paper proposes a novel method to diagnose sensory ataxia via an automated Romberg Test – the current de facto medical procedure used to diagnose this condition. It utilizes a convolutional neural network to predict joint locations, used for the calculation of various bio-mechanical markers such as the center of mass of the subject and various joint angles. This information is used in combination with data filtering techniques such as Kalman Filters, and center of mass analysis which helped make accurate inferences about the relative weight distribution in the lateral and anterior-posterior axes, and provide an objective, mathematically based diagnosis of this condition. In order to evaluate the performance of this method, testing was performed using dual weight scales and pre-annotated diagnosis videos taken from medical settings. These two methods both quantified the veritable weight distribution upon the ground surface with a ground truth and provided a real-world estimate of accuracy for the proposed method. A mean absolute error of $0.2912\%$ was found for the calculated relative weight distribution difference, and an accuracy of 83.33% was achieved on diagnoses.**

## I. INTRODUCTION

The Romberg Test (RT) was originally described by Marshall Hall, Moritz Romberg, and Bernardus Brach [1] to assess the integrity of the dorsal column pathway of the brain and spinal cord, responsible for regulating proprioception. Although commonly used, this test has several obvious flaws. These include the fact that the test has no objective criteria for a positive diagnosis except the practitioner's personal and subjective opinion, and the fact that many patients with sensory ataxia (SA) struggle to physically show up to a doctor's appointment for diagnosis. These drawbacks often hinder early diagnosis [2], when the condition is less obvious but still developing over time, as well as in the accessibility of the test for those with a more developed case of SA. To determine the objective criteria for a positive diagnosis through the RT, a research group in 2009 published a paper using electric analysis of the patients, suggesting that normal subjects deviate on average 6-7% of their body weight in the lateral direction, and 12-14% in the anterior-posterior direction [3]. This study provides the foundation for our proposed method, as it is the only concrete and quantifiable diagnosis criteria for SA available in the literature.

## II. BACKGROUND

SA is a very well-studied topic, and there have been a variety of attempts at developing methods to quantify it visually and diagnose it in the past. Previously used methods included the use of pressure sensors, visual markers, pixel-level illumination changes, and machine learning techniques.

### A. Visual Markers

Viet Hung Dao and Thi Thu Hien Hoang utilized two colored stickers placed on the subject's back in a vertical line to allow a camera and image processing program to infer necessary information [4]. This setup involved color filtering to detect the stickers, thresholding to extract the sticker's positions, and morphological processing to reduce noise in the binary image. Once the data was extracted from the image, contours were able to be identified which led to an accurate calculation of angles, corresponding to different levels of lean in the subject.

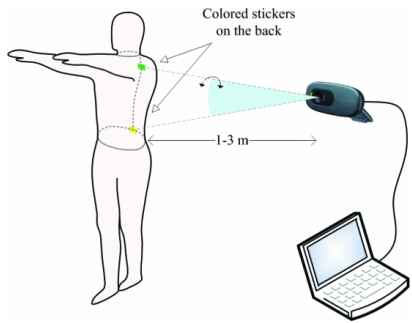

Figure 1: Proposed system setup to use colored stickers [4]

Although this method is effective in providing a low-cost and efficient way of calculating lean angle, the lean angle alone is not enough to provide a quantitative diagnosis for the RT: it does not provide information about the weight distribution of the subject. Another issue with this method is that placing stickers on the back of the subject is both inaccessible and prone to error (i.e. the stickers may not be placed perfectly in-line, creating bias in the measurement).

### B. Illumination Changes

A method to identify changes in posture was proposed by A. Nalci et al. which used an event camera (DVS128 from iniLabs) and image processing algorithms to measure illumination changes in key points around the body [5]. This was done by computing an unsteadiness value (U) as a function of total motion activity for each frame, inferred from illumination changes at key points. The output value from the program was compared with the ground truth data given by a balance board, measuring the center of pressure sway during balance tests. This method achieved a notable correlation of 0.98 – 0.99, although it still had certain drawbacks. One of the major drawbacks of this study was that it only involved a very small sample size of testing (10 adult participants, with just 3 of those 10 being female), and it is not yet proven to be effective in a clinical setting. Another drawback of this method is that it relies on the availability of an event camera – a type of camera that responds to local changes in illumination. These types of cameras are both very sparsely used and very expensive, and even perform surprisingly badly in low light conditions [6]. This factor makes this method less accessible for the public.

### C. Pose Estimation / Machine Learning

Machine learning was employed in a proposed method by N. Seo et al. which aimed to quantify the RT using pose estimation [7]. This method calculated the angles of four anatomical key points (neck, shoulders, pelvis, and knees) via a camera mounted at 180cm, and calculated the variance over time of these angles. Variance was compared between a control group and a group with SA to determine the relationship between SA and the variation of these angles. The Mann-Whitney U test was used to determine the statistical significance between the groups, which resulted in a p-value ≤0.001. This method shows promising results for the use of pose estimation for automating the RT, but it also has some flaws. For one, the method requires a camera to be mounted at exactly 180cm of height, which is unrealistic to achieve perfectly from home without the use specialist equipment such as a tripod. Additionally, the results of this method are never being evaluated against any ground truth such as a weight distribution linked to a positive or negative diagnosis of SA but are instead only being shown to identify a difference between people with SA and without. This might mean that when applied to the real world, edge cases that were not seen in the data may have a higher chance of being mis-classified.

### D. 3D Reconstruction

One method that uses a body reconstruction approach is mentioned in a paper by T. Kaichi et al. [8]. This method aimed to calculate the center of mass (CoM) of a human subject using a multi-camera setup, with each camera capturing a different angle of the human body to produce a 3D reconstruction of the body. Additionally, the body is segmented into nine different regions, with each region corresponding to a different weight contribution estimate. This approximation allowed the method to produce a reliable estimate of the CoM location within 10mm of error, and it works without any specialized equipment. Under the right conditions, this method was shown in the paper to be a very accurate and efficiently computed CoM estimate, however this was not the case when clothing was introduced, in particular loose-fitting clothing.

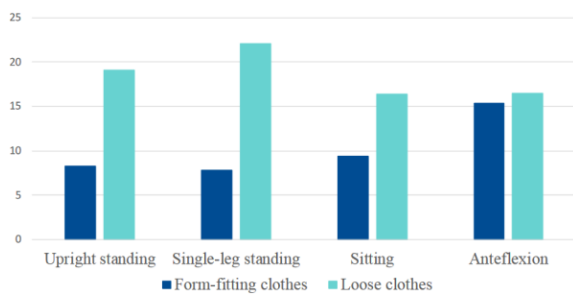

Figure 2: Comparison of CoM errors between subjects wearing loose-fitting clothing and form-fitting clothing [8]

This is a major limitation of this method, as variability in the clothing of individuals is natural. This is especially important in the context of our proposed method, as individuals with potential mobility impairments will be less comfortable altering their apparel to conduct this test.

*E. Summary of Limitations*

Upon reviewing related prior research, the following limitations were found to be the most common and influential in prior methods:

- Non-accessible equipment such as high-end specialized cameras being required.
- A lack of proven effectiveness against a ground truth
- Inability to adapt to different environments.
- A large room for error when using the method, as in the case of perfectly placed stickers being required.

III. PROPOSED METHOD

The proposed method aims to overcome the identified limitations of prior research by using image-processing techniques fed by just a basic phone camera, with no specialized equipment or setup needed. Important measurements were calculated frame-by-frame using OpenCV [9] to produce a reliable and accurate measurement of the weight distribution of the subject at all times to diagnose Sensory Ataxia via the Romberg Test. A video stream is captured from a front-on angle of the subject's complete body while the subject assumes the required position for the Romberg Test (feet together and hands by their side). This allows for accurate pose estimation and calculation of important angles and positions.

*A. Image Pre-Processing*

In our proposed method, image pre-processing was done before running any inference on the video to remove noise. This was done to reduce inaccurate joint predictions by the Convolutional Neural Network (CNN), and to improve the robustness of the method to variations in lighting conditions, image quality, and camera noise. Gaussian Blurring (GB) as described in [10] has been shown to improve the accuracy of CNNs from 0.9292 to 0.9750, a 5.3% increase in accuracy [11]. First, GB was applied to the raw frames of the video in order to reduce general noise. This technique involves convolving the image with a Gaussian Kernel, effectively applying a weighted average of each pixel's neighbor depending on distance. A kernel of size 5 was used in our method, as too high of a blur runs the risk of getting rid of too many details in the image. Additionally, a deep learning model from MediaPipe was used in order to identify regions of the image containing people. This model was used to output an image mask where every pixel not likely to be associated with a human subject would be heavily blurred using another Gaussian kernel of size 95. This blurring effectively draws attention from the Pose Detection CNN towards the subject, reducing the chance of a false positive and increasing the efficiency of the model.

A blur was chosen instead of simply removing the unnecessary pixels from the image, as it was observed that removing pixels had the effect of occasionally throwing off the pose detection.

*B. Joint Location Detection*

*1) BlazePose Pose Detection*

A CNN was employed to provide an accurate prediction of the 3D location of joints in the human body when given a raw image. This CNN named BlazePose was produced by Google Research[12] to address the problem of predicting body landmarks. It includes a set of thirty-three landmarks (seen below in Fig. 4) and was trained using a supervised learning approach on 85,000 pre-annotated. Due to detachment of detection and tracking modules (with detection only being run when needed), this model is very efficient when compared to competing models while maintaining accuracy. It was able to achieve 31 FPS on a basic Pixel 2 smartphone [12], which is stands out compared to OpenPose, a competitor pose-estimation model, which ran inference at 16 FPS on a desktop with a 12GB NVIDIA Xp GPU and i7-6800k CPU [13].

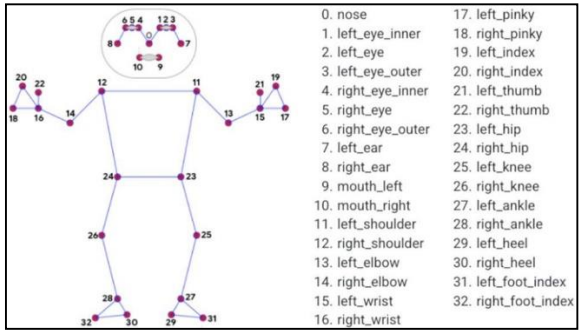

Figure 4: BlazePose Landmark Topology [12]

Using BlazePose means that the subject in our method will be able to be analyzed to a high degree of detail with a simple smartphone camera, while maintaining a low level of complexity and computing power needed to run inference over the video.

*2) Implementation*

When implementing and testing the pose estimation technique, certain joints (e.g feet, hands) were observed to be susceptible to noise, throwing off key calculations. This problem was solved by using a linear Kalman Filter (KF). KFs are powerful tools used for estimating the state of a dynamic system with noisy measurements. The key idea behind them is to recursively estimate the state of the system by combining a prediction of the state with the actual measurement [14]. They can be mathematically described by two main equations: the prediction step (1) and the update step (2).

$$\hat{x}_{k|k-1} = F_k \hat{x}_{k-1|k-1} + B_k u_k \quad (1)$$

$$P_{k|k-1} = F_k P_{k-1|k-1} F_k^T + Q_k \quad (2)$$

Where $\hat{x}_{k|k-1}$ is the predicted state at time $k$, $F_k$ is the state transition matrix describing how the state evolves over time in the system, $B_k$ is the control input matrix, $u_k$ is the control input, $P_{k|k-1}$ is the predicted state covariance matrix, and $Q_k$ is the process noise covariance matrix. This type of KF was chosen over other types (such as the extended or unscented KFs), as the movement of the foot joint was observed to follow a relatively straight trajectory, without any rotation or complex interactions with other joints. After applying this Kalman Filter, the inferred position of the joints followed the ground truth position to a significantly greater degree of precision, allowing for a much more stable and accurate estimate of the weight distribution.

*C. Centre of Mass (CoM) Calculation*

In order to calculate an accurate estimate of the position of the subject's CoM, the body was split into seven weighted regions (corresponding with a contribution to the mass of the body). The assignment of each body region was based on a study by P. Leva [15], shown below in Fig. 5.

| Segment | Endpoints Origin | Endpoints Other | Longitudinal length (mm) F | Longitudinal length (mm) M | Mass* (%) F§ | Mass* (%) M¶ |
|---|---|---|---|---|---|---|
| Head | VERT† | MIDG† | 200.2 | 203.3 | 6.68 | 6.94 |
| Trunk | SUPR† | MIDH‡ | 529.3 | 531.9 | 42.57 | 43.46 |
| UPT | SUPR† | XYPH† | 142.5 | 170.7 | 15.45 | 15.96 |
| MPT* | XYPH† | OMPH† | 205.3 | 215.5 | 14.65 | 16.33 |
| LPT | OMPH† | MIDH‡ | 181.5 | 145.7 | 12.47 | 11.17 |
| Upper arm | SJC‡ | EJC‡ | 275.1 | 281.7 | 2.55 | 2.71 |
| Forearm | EJC‡ | WJC‡ | 264.3 | 268.9 | 1.38 | 1.62 |
| Hand | WJC‡ | MET3† | 78.0 | 86.2 | 0.56 | 0.61 |
| Thigh | HJC‡ | KJC‡ | 368.5 | 422.2 | 14.78 | 14.16 |
| Shank | KJC‡ | LMAL† | 432.3 | 434.0 | 4.81 | 4.33 |
| Foot* | HEEL† | TTIP† | 228.3 | 258.1 | 1.29 | 1.37 |

Figure 5: Calculated percentage of contribution to total mass by each body segment [15]

According to this information, each joint located on the subject was grouped into one of these identified segments (ignoring MPT, Forearm, and Shank). From here, the calculation of the CoM is simply a weighted sum of the positions of each joint:

$$CoM = \sum_{i=1}^{n} w_i \cdot x_i$$

Where $n$ is the total number of joints, $x_i$ is the position of the $i$-th joint, and $w_i$ is the corresponding overall mass contribution of the $i$-th joint. See Fig. 6 for a demonstration of the inferred CoM location changing with motion.

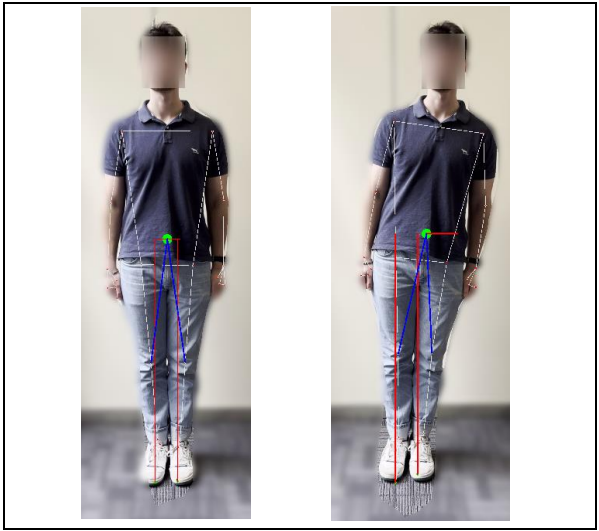

Figure 6: Estimated CoM shown as a green circle.

In order to smooth out sudden changes in the CoM due to jittering joint locations, our proposed method also uses an exponential moving average of the CoM location. This technique was chosen as it allows us to use a non-computationally intensive method to reduce jittering while providing more weight to recent observations of the joints:

$$x_t = \alpha \times y_t + (1 - \alpha) \times x_{t-1}$$

Where $x_t$ denotes the current smoothed CoM value, $y_t$ represents the current raw CoM value, $x_{t-1}$ is the previous smoothed CoM value, and $\alpha$ is the smoothing factor. The optimal $\alpha$ value of 0.9 was obtained through iterative experimentation based on observed noise compared with a ground truth (See section IV. (A)).

D. Relative Weight Distribution (RWD) Calculation

After defining the position of the CoM, calculations of the RWD on both axes became trivial using the positions of joints.

*1) Lateral Axis*

In order to calculate the RWD on the lateral axis, the horizontal distance between the CoM and the feet was used. Define $N_1, N_2$ to be the upward supporting forces for the right and left foot respectively, $m$ be the mass of the subject, and $g$ be the gravitational force on Earth ($9.81 ms^{-2}$), and $x_1, x_2$ be the perpendicular distances between the upward vector from the left and right feet respectively, and the CoM.

Net torque should be 0 ($Torque = Force \times Distance$):

$$N_1 x_1 = N_2 x_2$$

$$N_1 = N_2 \frac{x_2}{x_1}$$

Net force should be 0:

$$N_1 + N_2 = mg$$

$$N_2 \frac{x_2}{x_1} + N_2 = mg$$

$$N_2 = \frac{mg x_1}{x_1 + x_2}$$

$$N_1 = \frac{mg x_2}{x_1 + x_2}$$

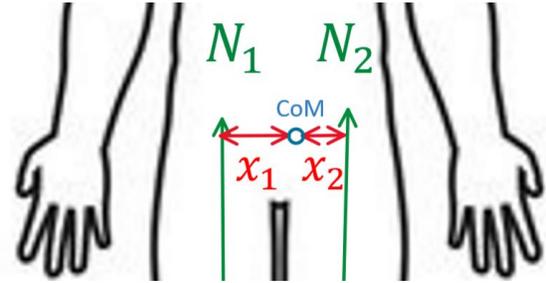

Figure 7: Diagram showing RWD calculation process for the Lateral Axis

Therefore, the relative weight distribution can be calculated as the absolute value of the difference between $N_1$ and $N_2$.

*2) Anterior-Posterior Axis*

As for the calculation of the RWD on the anterior-posterior axis, the vertical and horizontal distances between the ankle and CoM were calculated in order to obtain a right-angled triangle as shown below in Fig. 8. From here, the angle between the ankle and CoM was calculated as

$$\theta = tan^{-1}\left(\frac{d_h}{d_v}\right)$$

The horizontal component of the mass can then be calculated (where $W_{vert}$ is the weight of the subject):

$$W_{horiz} = W_{vert} \times tan(\theta)$$

This horizontal component is then interpreted as the ratio of weight on either side in the anterior-posterior axis.

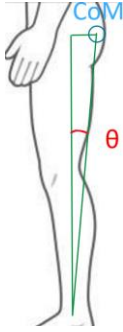

Figure 8: Diagram showing RWD calculation process for the Anterior-Posterior Axis

## IV. RESULTS

When testing, the method was executed on a PC running Linux Mint 21.2 Cinnamon, with an Intel® Core™ i7-9700 CPU @ 3.00GHz x 8 processor, and a NVIDIA GeForce RTX 2060 GPU. The method was developed using Python 3.10 in Visual Studio Code, heavily utilizing OpenCV-python v4.7.0.68. Additionally, the camera used to record testing data was an Apple iPhone 13 camera (12MP, 1920x1080 @ 30FPS).

The proposed method aimed to improve prior results in regard to the use of non-accessible equipment, the lack of ground truth to verify against, an inability to adapt to different conditions, and unrealistic expectations from the practitioner of the method in order to provide reliable results. For this reason, testing was done in two different manners:

1) Using a ground-truth method, with a dual-scale measure of weight bearing through the legs and assessing the numerical accuracy of the weight distribution calculations.
2) Using pre-annotated videos found online of patients with/without SA performing the Romberg Test with a practitioner and evaluating the proposed method's reliability in diagnosing these subjects.

### A. Dual-Scale Method

#### 1) Justification

This method was inspired by a study evaluating the use of the Dual-Scale Method for determining inequalities of hip bone density [16]. This method was calculated as having a Coefficient of Variation of 5.41%, meaning that the method is a reliable way to determine weight bearing through the legs.

#### 2) Testing

Testing was done via this method by instructing subjects to stand astride on two equally calibrated scales, giving a ground-truth weight measurement through each foot:

$$WD_R = \frac{S_R}{S_R + S_L} \qquad WD_L = \frac{S_L}{S_R + S_L}$$

Where $WD_R$, $WD_L$ is the weight distribution expressed as a ratio of the total weight on the right and left feet respectively, and $S_R, S_L$ is the scale reading for the right and left feet respectively. Subjects were instructed to lean at a number of differing angles to have a photo taken of both the scales and the subjects themselves, while keeping their feet flat on the ground. This testing method was performed with five different configurations based on the subject's sex, the viewing angle (forward or side), and clothing type (tight or loose). The manner of testing was also planned to have differing lighting conditions between photos. The results of this method of testing can be seen in the below table.

Table 1: Summary of Results

| Subject | RWD | Num. Photos | Mean Absolute Error |
|---|---|---|---|
| S1: Male, Front, Tight clothing | 0.46%-27.73% Mean: 9.63% SD: 7.29% | 18 | ±0.3294% |
| S2: Male, Side, Tight clothing | 1.16%-6.92% Mean: 3.80% SD: 2.04% | 14 | ±0.3086% |
| S3: Female, | 0.15%-6.68% | 13 | ±0.4615% |
| Front, Tight clothing | Mean: 3.25% SD: 2.90% | | |
| S4: Male, Front, Loose clothing | 2.08%-15.82% Mean: 7.19% SD: 4.15% | 17 | ±0.1165% |
| S5: Male, Side, Loose clothing | 0.23%-4.82% Mean: 2.27% SD: 1.67% | 10 | ±0.2400% |
| Combined | - | 72 | ±0.2912% |

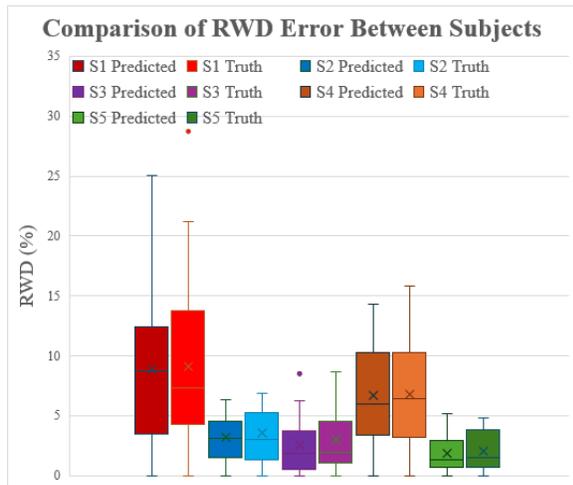

Figure 9: Boxplot showing differences between RWD error for each subject (grouped by color)

### B. Pre-Annotated Diagnosis Method

The other way that our proposed method was tested was using data available freely online. This method provided a way of getting data from a wide range of subjects in a clinical setting and gave the testing a real-world dimension for validation purposes. three SA-positive and three SA-negative videos were analyzed, all with a front-on view of the subject adhering to the requirements for the Romberg Test, and our proposed method returned a diagnosis after analyzing the video based on the maximum RWD observed. Our proposed method was able to correctly diagnose 2/3 of the SA-positive videos, and 3/3 of the SA-negative videos. This means that this method achieved a 100% accuracy on negative diagnoses, and a 67% accuracy on positive diagnoses, with an 83.33% accuracy overall.

### C. Discussion

As can be observed in Table 1, the results for the Dual-Scale method were relatively consistent, and all subjects achieved an MAE error <0.5%. Considering that the definition of a positive diagnosis of SA is said to be anywhere between 6-7% (1% margin of error) and anywhere between 12-14% difference (2% margin of error) in weight distribution for the lateral and anterior-posterior axes respectively, we consider our proposed method a success. Interestingly, the accuracy seemed to increase when loose clothing was worn instead of tight clothing. Additionally, the accuracy decreased for the female subject. An explanation for this could be that the employed CNN had less training data available for tight clothing and female subjects. It is also worth noting that females are on average shorter than males, and a shorter subject would increase the sensitivity of joint angle calculations, increasing potential error.

As for the low accuracy on diagnosing SA-positive subjects, it is worth noting that the incorrectly diagnosed subject was wearing black clothing with a black curtain as the background. While watching joint inferences in the video, it can be observed that the pose estimation consistently mis-predicted the location of the left knees and ankles of the subject. This had the effect of placing a bias on the estimated CoM throughout the analysis, causing a false-negative.

These results are an improvement on previous research into diagnosing SA. T. Kaichi et al. were not able to provide accurate measurements for the center of mass of a subject if they were to wear loose clothing, with an error of 19% [8] compared to our 0.1165% when the subject was to assume a standing position. A. Nalci et al. were able to achieve a higher accuracy of 98-99% on the center of pressure, however their method relied on a very expensive and inaccessible event camera, which would not be feasible for individuals to use at home. Additionally, although N. Seo et al. were able to achieve a notable p-value $\leq 0.001$ for the significance between calculated joint angles and SA diagnosis using machine learning, it is unclear whether this method could be applied well to a real-world scenario, as the only information available to the model was labels. Without such validation against real-world data, the reliability and generalizability of their findings remain uncertain.

### D. Limitations

Our proposed method has some limitations, especially around the testing performed. Although a satisfactory number of photos (72) were taken in a range of angles and lighting conditions, more insights could be gathered by introducing a more diverse range of subjects to the study. Additionally, although the Dual-Scale method was found to be effective and accurate [16], a more sophisticated ground-truth source such as the electric analysis method used by N. G Henriksoon et a.l [3] could be more appropriate at a higher financial cost.

Some issues were observed in the case where there was a bright white light source hitting a subject from the side, where it washed out their arm and caused the BlazePose model to clearly mis-predict the location of the elbow joint slightly. A similar phenomenon with identical clothing and background colors was also observed (as mentioned previously). Although these phenomena seem to be rare, they should be noted when setting up the RT environment. Further testing could be done in the future to further establish the likelihood of these phenomena and to quantify the impact that they have on the proposed method.

## V. CONCLUSION

The proposed method, using just a smartphone camera, achieved a mean absolute error (MAE) of just $\pm 0.2912\%$ with regard to the relative weight distribution. This is well within the required accuracy, as a positive Romberg Test was defined in a paper by N. G Henriksoon et. al. to be between 6-7% difference in weight distribution along the lateral axis, and 12-14% difference in weight distribution along the anterior-posterior axis (a 1% and 2% discrepancy allowance respectively). Additionally, the proposed method achieved an accuracy of 83.33% in the real-world through the pre-annotated diagnosis testing method.

These results are an improvement on previous related research using visual image processing. T. Kaichi et al. only achieved CoM estimation errors of 8% and 19% with tight- and loose-fitting clothes respectively. A. Nalci et al. required a method using a very expensive and inaccessible event camera, and N. Seo et al. were not able to test their method using a quantitative ground truth.

### A. *Future Research*

Some problems with the proposed method are that the testing was not extensive enough and lacks some sophisticated equipment which would increase the validity of results. Additionally, certain lighting conditions can make the method perform badly. Efforts in the future to resolve these problems could be directed at improving the testing process of the proposed method. In particular, a more diverse subject range would be beneficial in order to establish a more accurate performance and accuracy estimate on the entire population. Additionally, further study could be done to establish the impact of phenomena such as being washed out by a white light, and clothing seamlessly blending into the background when they share the same color palette.

Future developments with regards to capabilities with deep learning models should also be integrated into the proposed method in order to improve the accuracy and performance of segmentation and joint predictions. This would solve a lot of the current problems the method faces, and large strides in this area are currently taking place very often [17].